\documentclass{article}

\usepackage{arxiv}

\usepackage[utf8]{inputenc} % allow utf-8 input
\usepackage[T1]{fontenc}    % use 8-bit T1 fonts
\usepackage{hyperref}       % hyperlinks
\usepackage{url}            % simple URL typesetting
\usepackage{booktabs}       % professional-quality tables
\usepackage{amsfonts}       % blackboard math symbols
\usepackage{nicefrac}       % compact symbols for 1/2, etc.
\usepackage{microtype}      % microtypography
\usepackage{lipsum}
\usepackage{authblk}
\usepackage{wrapfig}
\usepackage{balance}
\usepackage[round, sort&compress, authoryear]{natbib}
\usepackage{float} 
\usepackage{times}
\usepackage{epsfig}
\usepackage{amsmath, amsfonts, amssymb}
\usepackage{bm}
\usepackage{algorithm}
\usepackage{algorithmicx}
\usepackage{comment}
\usepackage{enumitem}
\usepackage{fullpage}
\usepackage{tikz}
\usepackage{pifont}
\usepackage{amsthm}
\usepackage{subcaption}
\usepackage{mathtools}
\usepackage{multirow}
\usepackage[noend]{algpseudocode}
\usepackage{booktabs}
\usepackage[export]{adjustbox}
\usepackage{url}

\usepackage{xcolor}
\hypersetup{
    colorlinks=true,
    linkcolor=blue,
    citecolor=blue,
    urlcolor=blue
}

\usepackage[textsize=scriptsize,backgroundcolor=green]{todonotes}
\newtheorem*{remark}{Remark}

\newcommand\numberthis{\addtocounter{equation}{1}\tag{\theequation}}

\usepackage{xspace}
\newcommand*{\eg}{\textit{e.g.},\@\xspace}
\newcommand*{\ie}{\textit{i.e.},\@\xspace}
\newcommand*{\vs}{\textit{vs.}\@\xspace}
\newcommand*{\wrt}{\textit{w.r.t.}\@\xspace}
\makeatletter
\newcommand*{\etc}{%
	\@ifnextchar{.}%
	{\textit{etc}}%
	{\textit{etc.}\@\xspace}%
}
\makeatother
\algtext*{EndWhile}
\algtext*{EndFor}
\algtext*{EndIf}
\algdef{SE}[DOWHILE]{Do}{doWhile}{\algorithmicdo}[1]{\algorithmicwhile\ #1}%

\makeatletter
\def\BState{\State\hskip-\ALG@thistlm}
\makeatother

\title{Finding Pareto Trade-offs in Fair and Accurate Detection of Toxic Speech}

\author[1]{Soumyajit Gupta\textsuperscript{\dag}}
\author[2]{Venelin Kovatchev\textsuperscript{\dag}}
\author[3]{Anubrata Das}
\author[4]{Maria De-Arteaga} 
\author[3]{Matthew Lease}
\affil[1]{Dept.\ of Computer Science, The University of Texas at Austin}
\affil[2]{School of Computer Science, The University of Birmingham}
\affil[3]{School of Information, The University of Texas at Austin}
\affil[4]{McCombs School of Business, The University of Texas at Austin}
\affil[ ]{\texttt{\small 
$^1$smjtgupta@utexas.edu,
$^2$v.o.kovatchev@bham.ac.uk,
$^3$anubrata.das@utexas.edu,$^4$dearteaga@mccombs.utexas.edu
$^3$ml@utexas.edu}}

\begin{document}

\maketitle

{\let\thefootnote\relax\footnote{{\dag~contributed equally to this work.}}}

\begin{abstract}
    Optimizing NLP models for fairness poses many challenges. Lack of differentiable fairness measures prevents gradient-based loss training or requires surrogate losses that diverge from the true metric of interest. In addition, competing objectives (\eg accuracy \vs fairness) often require making trade-offs based on stakeholder preferences, but stakeholders may not know their preferences before seeing system performance under different trade-off settings. To address these challenges, we begin by formulating a differentiable version of a popular fairness measure, Accuracy Parity, to provide balanced accuracy across demographic groups. Next, we show how model-agnostic, {\em HyperNetwork} optimization can efficiently train arbitrary NLP model architectures to  learn {\em Pareto-optimal} trade-offs between competing metrics. Focusing on the task of toxic language detection, we show the generality and efficacy of our methods across two datasets, three neural architectures, and three fairness losses.
\end{abstract}

\section{Introduction}

Toxic language in social media is often associated with  various risks and harms: cyber bullying, discrimination, mental health, and even hate crimes. Given the massive volume of user-generated content online, manual review of all posts by human moderators simply does not scale. Consequently, natural language processing (NLP) methods have been developed to fully or partially automate toxicity detection \citep{schmidt-wiegand-2017-survey}. Prior work has achieved high Accuracy and F1 scores on toxicity detection (\eg \citep{zampieri-etal-2020-semeval}) across various model architectures: \eg convolutional (CNN) \citep{gamback2017using}, sequential (BiLSTM) \citep{graves2005bidirectional}, and transformer (BERT) \citep{devlin2018bert}. However, studies have also found that model accuracy can vary greatly across sensitive demographic attributes, such as race or gender \citep{park-etal-2018-reducing,sap2019risk,das2021fairness}. Subjective annotation in such tasks arise from personal biases and experiences of annotators. Traditional approaches relying on majority voting to resolve disagreements leads to oversimplification of the task. For example, a BERT-based classifier obtains $90.4\%$ \vs $84.5\%$ accuracy for White \vs African American author on Davidson's dataset \citep{davidson2017automated} when just optimized for overall accuracy, independent of author groups. Thus, in subjective domains, the minority viewpoint plays an important role \citep{sang2022origin}, where context and interpretations around data collections \citep{rahman2022understanding}, sources \citep{chaudhry2022you} and targets, can heavily influence judgments.

While recent years have seen rapid progress in fairness research, it is often measured in a post hoc manner, and optimization is often indirect (\eg by improving training data through pre- or post-processing) \citep{sap2019risk}. A particular challenge is that most existing measures are non-differentiable and thus cannot be optimized directly via gradient descent. While one can optimize differentiable surrogate loss functions instead, this risks {\em metric divergence} between the optimization criteria used in training \vs the actual metrics of interest
\citep{morgan2004direct,metzler2007linear,yue2007support,swezey2021pirank}.

As \citep{friedler2021possibility} and others have noted, different worldviews lead to conflicting definitions of fairness that are mutually incompatible and specific fairness measures must be selected (suitable to the given task, context, and stakeholders at hand). In this work, we adopt a popular fairness objective, Accuracy Parity \citep{zhao2020conditional} to optimize a model to provide balanced accuracy across demographic groups \citep{mitchell2021algorithmic,heidari2019moral,berk2021fairness,das2021fairness}. Because no differentiable version of this measure exists, we formulate a novel, differentiable version, \textit{Group Accuracy Parity} (GAP) that can be directly used to optimize descent-based models. We provide both a theoretical derivation and an empirical justification for GAP. 

However, optimizing GAP alone may reduce Overall Accuracy (OA) since seeking to better fit minority group may lead to worse fit of majority group that tends to drive OA. Ultimately, we face a trade-off between competing objectives,
% Moreover, such competing objectives abound in the world, as Friedler \citep{friedler2021possibility} highlights for fairness. 
whether we balance between competing accuracy goals (\eg precision \vs recall), fairness goals, or any combination thereof. Multi-Objective Optimization (MOO) provides a principled framework and rigorous toolbox for approaching such competing trade-offs, instead of treating them as single objective regularization problems \citep{soto2022application,little2023fairness,sorensen2024roadmap,suau2024whispering}. We believe such MOO work remains underexplored in NLP today, and to the best of our knowledge, ours is the first NLP work on MOO for fair toxic language detection.

Because competing objectives typically lack global optima, optimization requires choosing among a set of equally-valid, {\em Pareto-optimal} trade-offs between objectives. Naturally, selection of a suitable trade-off depends on stakeholder needs, and they typically wish to see system performance under real trade-off conditions before having to commit to any particular trade-off. We demonstrate how the full Pareto manifold -- for \textit{any} underlying model architecture -- can be efficiently induced, provided optimization can be performed via gradient descent (with differentiable loss objectives). This is accomplished via recent advances in {\em Pareto front learning} (PFL) \citep{navon2021learning,lin2020controllable,gupta2022scalable} for {\em HyperNetworks} \citep{ha2016hypernetworks}, which train one neural model to generate effective weights for a second, target model.

In summary, we pursue two distinct and complementary approaches for fair toxic language detection via model optimization. First, recognizing the repeated call for balancing accuracy across demographic groups, yet finding no differentiable metric doing so, we present the first differentiable version, GAP, enabling optimization for the first time via standard gradient descent. Our results show a clear benefit of optimizing directly for the target metric of interest rather than surrogate loss functions that diverge from it. Second, to demonstrate generality of PFL optimization over competing objectives, we induce the full Pareto front of optimal trade-offs between OA \vs three different fairness measures: GAP and two prior measures. 
To show generality of both techniques -- single-objective GAP and multi-objective PFL -- we show optimization over three distinct neural architectures (CNN, BiLSTM, and BERT) on two datasets: Davidson \citep{davidson2017automated} and Wilds \citep{koh2021wilds}. 

Our results show that GAP better balances accuracy across demographic groups (authors and targets of potentially toxic tweets) than existing differentiable measures. With multi-objective PFL, we show that we can successfully induce the full manifold of Pareto-optimal trade-offs across all differentiable objectives and neural architectures considered. GAP also achieves the best empirical trade-offs for OA \vs balanced accuracy in comparison to the two other fairness metrics considered. Finally, we note that GAP and PFL are broadly applicable and can be adapted for a wide range of NLP tasks, beyond the task of toxicity detection. For reproducibility and adoption, we provide our GAP source code.\footnote{\tt{Source code at \url{https://github.com/smjtgupta/GAP}.}}.

\section{Related Work}

\subsection{Toxic Language Detection and Fairness}

Many datasets now exist to train and test automated systems for TL detection \citep{poletto2021resources,vidgen2020directions}. Many NLP models have been proposed and continue to increase overall accuracy of detection \citep{fortuna2018survey,macavaney2019hate,schmidt-wiegand-2017-survey}. However, recent studies highlight the racial bias induced in such classification tasks. \citep{davidson2017automated} introduced a dataset with a corpus of tweets collected from social media and human annotations on the toxicity of the tweet. \citep{sap2019risk} and \citep{davidson2019racial} analyze the correlation between race and gold-label of toxicity in the \citep{davidson2017automated} dataset and find a strong association between AAE markers and toxicity annotation, where both of the works noisily infer author dialect via  \citep{blodgett2017dataset}'s model as a proxy for race. The Wilds \citep{koh2021wilds} dataset contains targets of TL with different demographic information and human annotated majority voted labels. It provides predefined training/test splits for effectively measuring distribution shifts in TL models. To address the problem of bias in automatic TL detection, some work has been done on improving the training and testing data \citep{park-etal-2018-reducing,sap2019risk,rottger-etal-2021-hatecheck}, with the expectation that fairer data will lead to fairer learned models. Some of the most similar to our work, by \citep{xia2020demoting}, \citep{10.1145/3442188.3445875}, and \citep{shen2022optimising}, seeks to reduce the bias towards AAE-authors in the algorithm rather than data.

\subsection{Fairness Measures} \label{sec:fairness}

The amplification of systemic unfairness through AI applications has been pronounced across different critical application areas such as hiring, finance, legal applications, content moderation \etc \citep{angwin2016machine,balashankar2022need}. It is of societal and ethical importance to examine if an AI is discriminative and develop methods to make the AI fair on grounds such as gender, ethnicity, or other forms of identity attributes \citep{ekstrand2019fairness}. To connect fairness concepts with statistical measures in machine learning, \citep{mitchell2021algorithmic} synthesizes fairness measures based on the confusion matrix. \citep{friedler2019comparative} further categorize fairness measures into largely three categories: 1) measures based on base rates, such as Disparate Impact \citep{feldman2015certifying}, 2) measures based on group-conditioned accuracy, and 3) measures based on group-conditioned calibration.

\subsection{Pareto Optimization of Trade-offs}
\label{sec:rw_pareto}

Multi-Objective Optimization (MOO) is increasingly pursued in fair classification \citep{caton2020fairness}. The complexity of real-world problem often leads to competing objectives such as accuracy \vs fairness. Pareto frameworks are powerful tools to balance between such competing objectives. Several works \citep{balashankar2019fair,martinez2020minimax} seek to balance accuracy \vs fairness. \citep{valdivia2020fair} presents a group-fairness based trade-off model for decision tree classifiers via a genetic algorithm. % NSGA-II.
%which has the same convergence and reproducibility issues mentioned earlier. In addition, their reported results violate fundamental definitions of Pareto optimality. 
\citep{wei2020fairness} uses Chebyshev scalarization to provide a neural architecture for fairness \vs accuracy Pareto front computation in classification. 
%, which assumes that objective functions must sum up to a constant.
%but do not justify this assumption. 
\citep{linx2019pareto} claims Pareto optimality on the basis of KKT conditions. In this work, we adopt \citep{gupta2022scalable}'s SUHNPF framework, given its error tolerance bounds and strong empirical performance. We apply it as a HyperNetwork \citep{ha2016hypernetworks} 
%as it is capable of handling both non-convex functions and constraints, a feature that most current MTL solvers are lacking. The capabilities and efficiency of SUHNPF makes it the preferred framework 
to optimize a variety of neural network models for TL detection. While we only optimize the Pareto tradeoff between a single accuracy measure \vs a single fairness measure, the framework itself is more general and directly supports optimizing arbitrary numbers of competing objectives (and constraints).

\section{Group Accuracy Parity (GAP)}\label{sec:GAP}

In this work, we focus on {\em Accuracy Parity} (AP) \citep{zhao2020conditional}, \ie balancing accuracy across groups (sub-populations based on some demographic criteria), sometimes known as {\em equal accuracy} \citep{mitchell2021algorithmic}, {\em equality of accuracy} \citep{heidari2019moral}, {\em overall accuracy equality} \citep{berk2021fairness}, {\em accuracy equity} \citep{dieterich2016compas}, or {\em accuracy difference} \citep{das2021fairness}. We do not claim any primacy of this particular notion of fairness, but show that if one is interested in it, it can be directly optimized via our Group Accuracy Parity (GAP) measure without {\em metric divergence}  \citep{morgan2004direct,metzler2007linear,yue2007support,swezey2021pirank} between loss function \vs evaluation metric.

\subsection{Accuracy Difference}

While AP is an equality condition, we still need to quantify the deviation from equality in cases of unequal performance across groups. We therefore use {\em Accuracy difference} (AD) \citep{das2021fairness}, a continuous version of AP to measure this deviation. AD is shown in (\textbf{Eq. \ref{eq:ad}}), where $\hat{y},y,g$ are the predicted label, true label, and group attribute respectively.
\begin{align}
	% \resizebox{0.42\textwidth}{!}{
		AD =  \underbrace{P[\hat{y}=y|g=1]}_\text{Acc Group 1 (g=1)} - \underbrace{P[\hat{y}=y|g=0]}_\text{Acc Group 0 (g=0)} \label{eq:ad}
	% }
\end{align}

AD being defined based on the confusion matrix, makes the formulation is probabilistic in nature, \ie ratio of numbers over the dataset, and not distribution over variable, AD becomes non-differentiable. Thus, AD can only be used in a post-hoc manner and cannot be directly used for gradient-based back propagation. Furthermore, Eq. \ref{eq:ad} inherently assumes that the majority group accuracy $(g=1)$ will always be higher than the minority group $(g=0)$, which might not always hold true, resulting in potential negative values of AD in the range [-1,1]. Naturally, as a post-hoc measure, AD is disconnected from the optimization objective of the model used during training.
 
These limitations motivated us to define a differentiable, non-probabilistic form of AD we refer to as {\em Group Accuracy Parity} (GAP), which allows any descent-based model during training to optimize close to equal accuracy across sensitive attribute classes, and addresses the range issue of AD.

% \vspace{-0.5em}
\subsection{Formulation}

Binary Cross Entropy (BCE), as formulated in  \textbf{Eq. \ref{eq:bce}} is typically used as a loss function for optimizing a classifier. Although not a strict one-to-one correspondence, it is observed that minimizing BCE leads to maximization of Accuracy. 

\begin{align}
	% \resizebox{0.42\textwidth}{!}{
		BCE = -\frac{1}{N} \sum \limits_{N} y \log (\hat{y}) + (1-y) \log (1-\hat{y})
  \label{eq:bce}
	% }
\end{align}
\vspace{-1.0em}

Weighted Cross Entropy (WCE) is a variant of BCE that re-weights the error for the different classes proportional to their inverse frequency of labels in the data. The class re-weighting strategy is available in packages like SkLearn \citep{scikit-learn} and is discussed in detail by \citep{lin2017focal}. For balanced classification across sensitive attributes (\eg demographic information across author groups or gender information across targets in Hate Speech), we formulate our GAP loss function as follows: we first calculate the WCE for each sensitive attribute ($g$), then minimize the difference across them. The GAP loss function in \textbf{Eq. \ref{eq:GAPad}} is minimized only when WCE errors match across the binary sensitive attribute. 

\begin{align}
	% \resizebox{0.42\textwidth}{!}{
		GAP = \underbrace{WCE_{overall}}_\text{Overall Acc} \,\, + \,\, \lambda \, \| \underbrace{WCE(g=1)}_\text{Acc Group 1 (g=1)} - \underbrace{WCE(g=0)}_\text{Acc Group 0 (g=0)}\|_2^2 \label{eq:GAPad}
\end{align}

\noindent The GAP function has the following properties: 

\begin{enumerate}
    \item \textbf{GAP maps to AD.} GAP has a one-to-one correspondence to AD \ie minimizing GAP also minimizes AD.
    \item \textbf{GAP is differentiable.} GAP is defined as the squared 2-norm difference between the Weighted Cross Entropy (WCE) across the two sensitive attribute. Since WCE is differentiable, so is the 2-norm difference. Hence GAP can optimize any descent based model. 
    \item \textbf{GAP is symmetric.} GAP has a 2-norm formulation, ensuring the range of attainable values are within $GAP \in [0,1]$, avoiding the negativity issue faced in AD. Also being a 2-norm measure, the loss surface of GAP is smoother than other comparable measures like CLA \citep{shen2022optimising}, which uses 1-norm \citep{boyd2004convex}.
\end{enumerate}

For a step-by-step derivation from WCE to GAP, readers are referred to \textbf{Appendix A}, showing the strict correspondence between the loss measures. In this paper we implement GAP (Eq. \ref{eq:GAPad}) to correspond to AD  (Eq. \ref{eq:ad}). As such, GAP can be optimized over binary labels and binary groups.

% \vspace{-0.5em}
\section{Optimizing Competing Objectives} \label{sec:optimize}

Typically, toxicity detection systems are trained with the single objective of maximizing OA  \citep{Founta18,park-etal-2018-reducing,rottger-etal-2021-hatecheck} or a custom defined objective \citep{xia2020demoting}. In contrast, we frame toxicity detection as a Multi-Objective Optimization (MOO) problem. It is important to highlight the distinction between an M(Multi)OO \vs S(Single)OO  formulation and their interpretation. Consider the two objectives as $f_1$: Cross-Entropy and $f_2$: Fairness. Traditional fair classifiers operate by adding a penalty term corresponding to Fairness to the main objective Entropy with a hyper-parameter $\lambda$ in \textbf{Eq. \ref{eq:hatesoo}}.\begin{align*}
    &\underset{}{\min} \quad f_1 + \lambda f_2 \\
    &\underset{}{\min} \quad \textrm{Cross-Entropy loss} + \lambda \, \textrm{Fairness loss} \numberthis \label{eq:hatesoo}
\end{align*}

The reader is specifically requested to note that such optimization process does not have any control over the range of $\lambda$, and it can vary generally between $(0, \infty)$. During the optimization process, we tune $\lambda$ till we get a desired performance in SOO setting. Furthermore, there is no explicit requirement of the scale of $f_1$ and $f_2$ to be the same. Thus, there is no simple correlation between the the amount of Fairness we want \vs the value of $\lambda$.

An unconstrained MOO problem with two competing loss objectives is defined in \textbf{Eq. \ref{eq:hatemoo}}. Note that this is a joint min-min problem instead of a single min problem. The objectives here need to be at the same scale \wrt each other. If the expectation is to achieve a liner trade-off between them,  the linear scalarized form of the MOO problem with trade-off $\alpha \in [0,1]$, minimizes both objectives simultaneously in \textbf{Eq. \ref{eq:hatemooscal}}. Solving this reformulated MOO problem would achieve balance between Entropy and Fairness, with $\alpha$ holding strict mathematical interpretation of linear trade-off. % across sensitive attribute of the classifier.
Decreasing Entropy causes Fairness to increase, while decreasing Fairness causes Entropy to increase.
\vspace{-0.5em}

\begin{align*}
    &\underset{}{\min \min} \quad f_1 \, , \, f_2 \numberthis \label{eq:hatemoo} \\
	&\underset{}{\min} \quad \alpha f_1 + (1-\alpha) f_2 \\
	&\underset{}{\min} \quad \alpha \, \textrm{Cross-Entropy loss} + (1-\alpha) \, \textrm{Fairness loss} \numberthis \label{eq:hatemooscal}
\end{align*}

Note that there are multiple mathematically optimal solutions to \textbf{Eq. \ref{eq:hatemooscal}}. Every optimal solution corresponding to each value of $\alpha$ in Eq. \ref{eq:hatemooscal} is a member of the Pareto optimal solution set \ie the Pareto front contains the set of optimal model parameters given the dataset and the model. To solve this MOO problem, we adopt the SUHNPF Pareto framework \citep{gupta2022scalable} as a HyperNetwork \citep{ha2016hypernetworks} to learn optimal TL detection neural model parameters over trade-offs. Hypernetworks train one neural model to generate effective weights for a second, target model. 

SUHNPF efficiently learns the entire Pareto manifold of feasible trade-off values during training. This empowers users to then choose any solution point they prefer on the manifold, \textit{a posteriori}, and extract the classifier weights configuration as per their desired trade-off $\alpha$, without retraining the model for that $\alpha$. Training the same model for $K$ different $\alpha$'s, with $R$ being the time for a single run, would result in total runtime of $K \times R$ \ie linear on the number of runs. Using the Hypernetwork to learn the manifold is computationally much more efficient \ie taking a constant time $c \times R$, $1 < c \ll K$ over feasible $\alpha$'s, rather than for each value of $\alpha$. Refer to \textbf{Appendix \ref{app:runtime}} for values on runs.

%\clearpage
\vspace{-0.5em}
\section{Experimental Details}

In this section, we describe our datasets, neural models, baseline losses and other evaluation details.

% \subsection{Setup}

% Experiments use a Nvidia 2060 RTX Super 8GB GPU, Intel Core i7-9700F 3.0GHz 8-core CPU and 16GB DDR4 memory. We use the Keras \citepp{chollet2015} library on a Tensorflow 2.0 backend with Python 3.7 to train the networks in this paper. For optimization, we use AdaMax \citepp{kingma2014adam} with parameters (\textit{lr}=0.001) and $1000$ steps per epoch. 
%See additional details in \textbf{Appendix \ref{app:implement}}. 

\subsection{Datasets}

We consider two datasets: \citep{davidson2017automated} for author demographics and the {\em Civil Comments} \citep{borkan2019nuanced} portion of {\em Wilds}
\citep{koh2021wilds} for target demographics (\textbf{Table \ref{tab:data_specs}}). In each case, we frame the task as a binary classification problem (Toxic \vs non-Toxic, or ``safe'') with binary sensitive attributes (Majority \vs Minority, the under-represented,  sensitive attribute). Note that ``Majority'' and ``Minority'' in our work simply refers to the statistical representation of the group in the data and does not carry any social or cultural meaning.

\begin{table}[ht]
	\centering
	% \vspace{-0.5em}
	% \resizebox{\linewidth}{!}{%
	\begin{tabular}{c|c|rrr}\toprule
		Dataset & Group & Toxic & Safe & Total\\ \midrule
		\multirow{2}{*}{Davidson} & Minority & 8,725 & 302 & 9,027 (36\%)\\
		& Majority & 11,895 & 3,861 & 15,756 (64\%)\\ \midrule
		\multirow{2}{*}{Wilds} & Minority & 5,973 & 33,762 & 39,735 (44\%)\\
		& Majority & 6,832 & 42,950 & 49,782 (56\%)\\ \bottomrule
	\end{tabular}
        % }
	% \vspace{-0.5em}
	\caption{Statistics of the two datasets used in this work. For \citep{davidson2017automated}, we consider the author demographics AAE \vs SAE as group attribute for minority \vs majority group attributes. For Wilds \citep{koh2021wilds}, we consider the binary group target gender as male \vs female for minority \vs majority group attributes.}
	\vspace{-1em}
	\label{tab:data_specs}
\end{table}

\textbf{Author Demographics Dataset} We consider fair moderation of posts written by authors from different demographic groups in \citep{davidson2017automated}. Prior studies \citep{sap2019risk,arango2019hate} have empirically demonstrated the existence of bias towards author demographics in toxic language classification. The sensitive attribute in this dataset is \textit{race}, as identified by the dialect of the tweets. Following prior work, we apply \citep{blodgett2017dataset}'s model to automatically-detect dialect labels for each of the tweet as African-American English (AAE) or Standard American English (SAE), representing \textit{Minority} and \textit{Majority} groups, respectively. We acknowledge both that dialect is only a weak surrogate representation of demographic race, and that automatic detection of dialect will naturally incur noise. However, in this, we follow established practices from prior work. Our fairness methods are agnostic to the sensitive attribute labeled in the data, and our results are only intended to attest to the capabilities of our proposed methods, rather than provide findings regarding protection of any specific vulnerable population. 
\citep{davidson2017automated}'s data includes 24,783 Twitter posts labeled as Hate, Offensive, or Normal. Following prior work \citep{park-etal-2018-reducing}, we 
%are framing the problem as a binary classification task, by setting class
set the class label to 1 (Toxic) if the post contains hate speech or offensive language, and 0 otherwise. 
We note that tweets from \textit{Minority} authors are annotated as toxic in 96\% of the cases, compared to 75\% for the tweets by \textit{Majority} authors. While these statistics suggest an important risk of annotation bias in this dataset, {\em dataset debiasing lies beyond the scope of our work}. Our focus in this work is restricted to balancing accuracy across the groups, given the dataset as it is annotated.

\textbf{Target Identity Dataset} To assess fair protection of different groups targeted in posts, we use the {\em Civil Comments} \citep{borkan2019nuanced} portion of {\em Wilds}
\citep{koh2021wilds}. %The modified dataset has standardized pre-specified splits. 
This dataset has 448,000 training tweets labeled as Toxic or non-Toxic. Each tweet has explicit annotation for the demographics, gender, or religion of the target entity. 
We select tweets where more than 50\% of annotators agreed on the gender of the target. %\textbf{Table \ref{tab:data_specs}} shows the distribution of targets and labels in the dataset
In this work, we include only female (majority) and male (minority) genders in order construct a binary sensitive attribute for our experiments. 
%\vk{female is the majority meaning that women are more often the TARGET of hate. Should we mention something about this?} 
In doing so, we fully-acknowledge both the non-binary nature of gender and individual freedom of self-identification. As noted above, our methods are agnostic as to the sensitive attribute labeled in the data, and our inclusion of only two genders merely reflects a convenient way to assess the capabilities of our proposed methods in regard to balancing accuracy across a binary sensitive attribute. 

\subsection{Neural Models Considered}

To assess the generality of our methods across distinct neural architectures, we evaluate over three types of models: CNN \citep{gamback2017using}, BiLSTM \citep{graves2005bidirectional} and BERT \citep{devlin2018bert}. For full experimental setup, please refer to \textbf{Appendix \ref{app:setup}}. For all three models, we freeze the feature representation layers and optimize the weights of the classification layer. In general, GAP loss optimization and the SUHNPF hypernetwork \citep{gupta2022scalable} support such generalization across any models that can be trained via gradient descent.

\subsection{Baseline Loss Functions}
\label{sec:adv_cla}

We compare against two baseline loss functions. The first fairness loss CLAss-wise equal opportunity (CLA) \citep{shen2022optimising} seeks to balance False Negative Rate (FNR) across protected groups \citep{chouldechova2017fair}, also known as equality of opportunity \citep{hardt2016equality}. CLA minimizes the error in absolute differences between error \wrt a label $(BCE(y))$ and error \wrt a label given the sensitive attribute $(BCE(y,g))$, with hyperparameter $\lambda \in [0,\infty]$, which differs from minimizing AD. Due to the 1 norm nature of CLA, the optimization surface for the loss function is not smooth \citep{boyd2004convex}. 

\begin{align}
	% \resizebox{0.42\textwidth}{!}{
		CLA = BCE + \lambda \cdot \sum_{y \in C} \sum_{g \in G} |BCE(y,g) - BCE(y)|  \label{eq:cla}
	% }
\end{align}

The second fairness loss \citep{xia2020demoting} is an adversarial approach to demoting unfairness, which we denote as ADV. It seeks to provide false positive rate (FPR) balance \citep{chouldechova2017fair} across groups, otherwise known as {\em predictive equality}.
Being adversarial in nature, this method and others \citep{chen2024hate} does not have any correspondence to any evaluation measure. Thus, users should take caution of possible metric divergence while using such techniques, with tuner $\beta \in [0,1]$.

\begin{align}
	% \resizebox{0.42\textwidth}{!}{
		ADV = \beta \cdot BCE + (1-\beta) \cdot (adversary(y,g)-0.5) \label{eq:adv}
	% }
\end{align}

However, while ADV is motivated by FPR balance, no equivalence between the loss function and the evaluation metric is shown, exemplifying {\em metric divergence} between loss function and evaluation goal. Their reported results also show only limited empirical correspondence between reducing the model loss and reducing FPR.

\subsection{Experimental Setup}

We have two experimental setups with the Weighted Cross Entropy (WCE) as $f_1$ and the Fairness criteria as $f_2$. First, we optimize the fairness measure directly as a SOO problem following Eq. \ref{eq:hatesoo} under a penalization setting, as proposed in CLA \citep{shen2022optimising}. Secondly, we use the MOO setting to find the best trade-offs between WCE and fairness measure following Eq. \ref{eq:hatemooscal}, with the SOO \vs MOO distinction described in \textbf{Sec \ref{sec:optimize}}. 
% \vk{we probably want to make a clear separation between the two experimental setups (with and without MOO). We should compare against baselines separately for each condition.}

\vspace{-0.5em}
\subsection{Evaluation Measures}

Our focus in this work is the tension between minimizing {\em accuracy difference} (AD) \citep{das2021fairness} and maximizing overall accuracy (OA). We thus evaluate on four post-hoc measures: OA over the dataset (majority and minority groups together), accuracy of each group separately, and AD observed between groups. Although we do not directly optimize F1, since a differentiable version of F1 does not exist, we still report the values in \textbf{Appendix \ref{app:diverge}}.

\vspace{-0.5em}
\section{Results}

\subsection{Existing Bias in CNN, BiLSTM, BERT}

\textbf{Table \ref{tab:baseline-david}} presents results for three toxic language classifiers optimized to maximize OA (\ie WCE) on \citep{davidson2017automated}'s dataset. The \textit{Majority} class consistently shows 6-7\% higher accuracy than the \textit{Minority} class, across models and five random initialization. Such imbalance serves as motivation for our work to optimize OA/AD across demographic groups. This unequal behavior in toxic language detection is consistent across all three neural models and both datasets. Due to space restrictions in the main body, we present the results only for the BERT-based classifier. However, our findings also apply to BiLSTM and CNN networks, whose results are available in \textbf{Appendix \ref{app:modelwilds}}.

\begin{table}[h]
	\centering
	% \vspace{-1em}
	% \resizebox{\linewidth}{!}{%
		\begin{tabular}{l|c|c|c|c}
			\toprule
			Models & Overall \% & Majority \% & Minority \% & AD \% \\ \midrule
			CNN & 87.52 $\pm$ 0.3 & 89.12 $\pm$ 0.2 & 82.88 $\pm$ 0.3 & 6.24 $\pm$ 0.2 \\
			BiLSTM & 87.60 $\pm$ 0.2 & 89.37 $\pm$ 0.2 & 82.46 $\pm$ 0.1 & 6.91 $\pm$ 0.3 \\
			BERT & 88.84 $\pm$ 0.2 & 90.35 $\pm$ 0.2 & 84.47 $\pm$ 0.1 & 5.88 $\pm$ 0.1 \\ \bottomrule
	\end{tabular}
 % }
	% \vspace{-0.5em}
	\caption{Baseline accuracy results on \citep{davidson2017automated}'s dataset %(over test split) 
	when maximizing overall accuracy (OA) only. Results show consistent bias of higher accuracy for the Majority.}
	\vspace{-1.0em}
	\label{tab:baseline-david}
\end{table}

\subsection{Single Objective Optimization (SOO)}\label{subs:res-soo}

\begin{table}[ht]
	\centering
	% \vspace{-2em}
	% \resizebox{\linewidth}{!}{%
		\begin{tabular}{l|cccc} \toprule
			Measure & Overall \% & Majority \% & Minority \% & AD \% \\ \midrule
			\multicolumn{5}{c}{Davidson} \\ \midrule
            Baseline & 88.84 $\pm$ 0.2 & 90.35 $\pm$ 0.2 & 84.47 $\pm$ 0.1 & 5.88 $\pm$ 0.1 \\
			\textbf{GAP (Ours)} & 87.32 $\pm$ 0.1 & 87.35 $\pm$ 0.1 & 87.26 $\pm$ 0.1 & \textbf{0.09 $\pm$ 0.0} \\
			CLA & 87.57 $\pm$ 0.2 & 87.82 $\pm$ 0.1 & 86.87 $\pm$ 0.1 & 0.95 $\pm$ 0.0 \\
			ADV & 86.27 $\pm$ 0.4 & 86.88 $\pm$ 0.2 & 84.52 $\pm$ 0.3 & 2.36 $\pm$ 0.1 \\ \midrule
			\multicolumn{5}{c}{Wilds} \\ \midrule
            Baseline & 84.68 $\pm$ 0.3 & 86.41 $\pm$ 0.2 & 82.49 $\pm$ 0.1 & 3.88 $\pm$ 0.2 \\
			\textbf{GAP (Ours)} & 84.38 $\pm$ 0.1 & 84.51 $\pm$ 0.1 & 84.23 $\pm$ 0.0 & \textbf{0.28 $\pm$ 0.0} \\
			CLA & 84.43 $\pm$ 0.1 & 85.23 $\pm$ 0.1 & 83.41 $\pm$ 0.0 & 1.82 $\pm$ 0.1 \\
			ADV & 83.61 $\pm$ 0.2 & 84.17 $\pm$ 0.1 & 82.91 $\pm$ 0.1 & 1.26 $\pm$ 0.1
			\\ \bottomrule
		\end{tabular}
	% }
	% \vspace{-1em}
	\caption{Optimizing fairness in a SOO setup. We compare a BERT-based model trained using cross entropy (Baseline) with models trained using different fairness measures. Our proposed measure (GAP) obtains the best results in reducing AD while maintaining high overall accuracy.}
	\label{tab:david_measure}
	\vspace{-1.0em}
\end{table}

\textbf{Table~\ref{tab:david_measure}} shows the results for the SOO experimental setup. The baseline BERT model optimized via Cross Entropy obtains 88.84\% OA and 5.88\% AD on \citep{davidson2017automated} and 84.68\% OA and 3.88\% AD on Wilds \citep{koh2021wilds}. All three loss functions successfully reduce the AD on both datasets. As expected, the improvement in fairness comes at the cost of lower OA. We evaluate the different optimization metrics by looking at both the change in AD and in OA.

ADV performs the worst of the three measures, most notably due to its relatively large drop in OA. Optimizing for GAP and CLA gives the same OA, where the two losses show no significant difference across 5 initialization. However, in terms of reducing AD, our GAP measure outperforms CLA by 0.9\% on Davidson and 1.5\% on Wilds. Looking at the results, we can conclude that GAP is the best performing measure in terms of reducing Accuracy Difference. The results are consistent across both datasets. These results show the value in optimizing a measure that correctly reflects the desired notion of fairness, as well as the benefit from directly optimizing the measure of interest, rather than surrogate or approximate loss functions, to avoid metric divergence.

\begin{figure}[h!]
\centering
\vspace{-1em}
    \begin{subfigure}{0.45\linewidth}
		\centering
		\includegraphics[width=\linewidth]{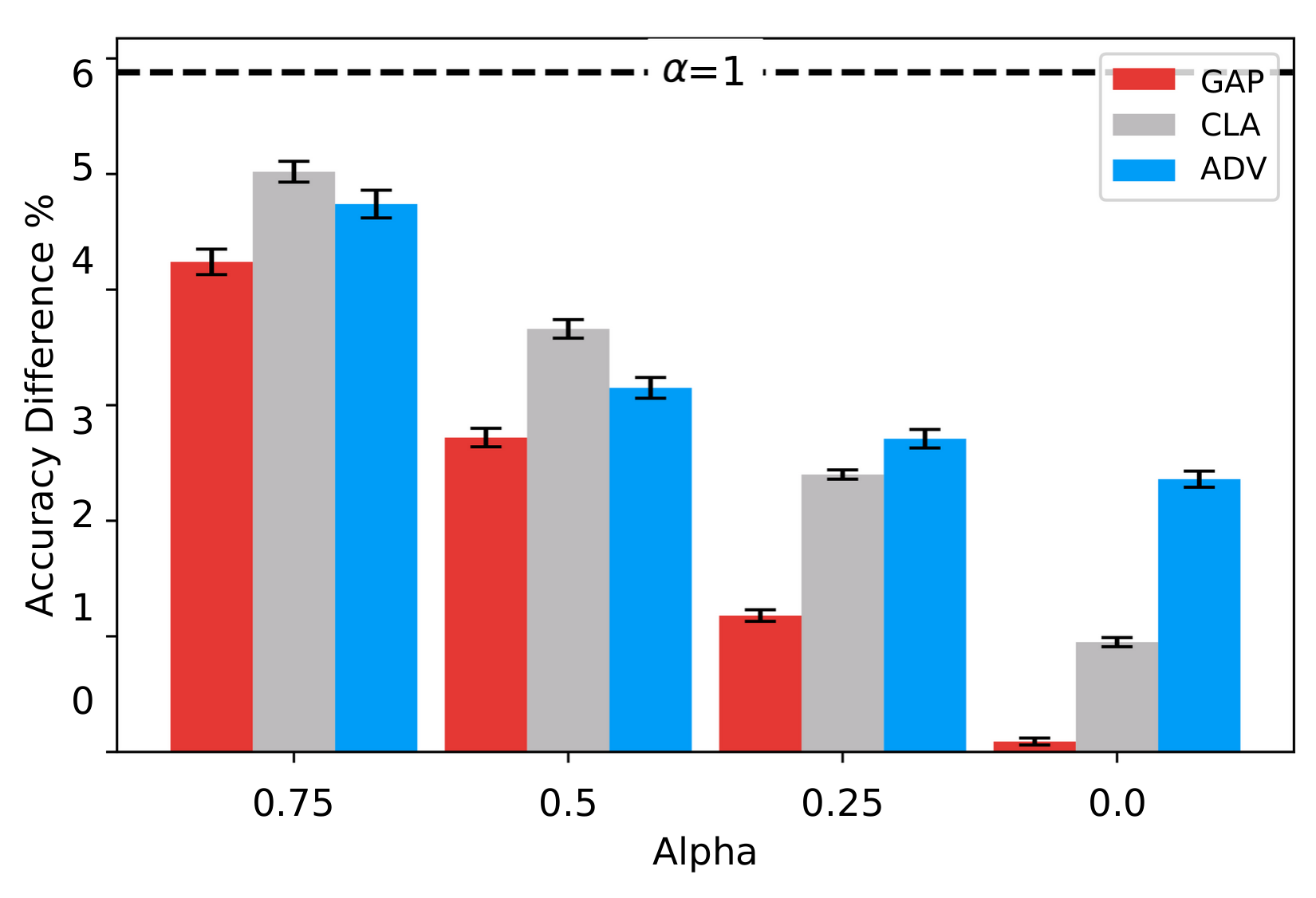}
		\vspace{-1.75em}
		\caption{Davidson: AD}
	\end{subfigure}
	\begin{subfigure}{0.45\linewidth}
		\centering
		\includegraphics[width=\linewidth]{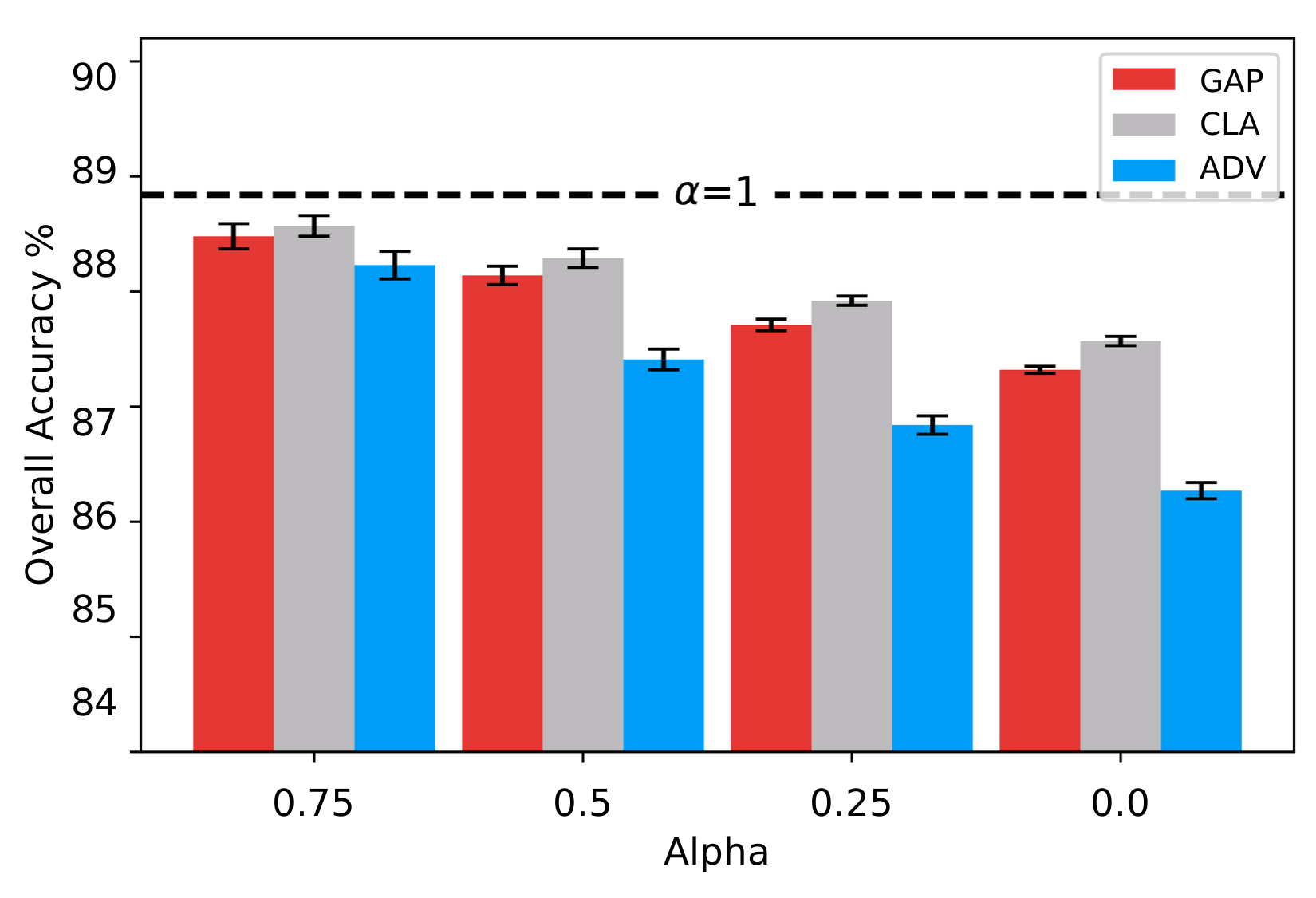}
		\vspace{-1.78em}
		\caption{Davidson: OA}
	\end{subfigure}
	\begin{subfigure}{0.45\linewidth}
		\centering
		\includegraphics[width=\linewidth]{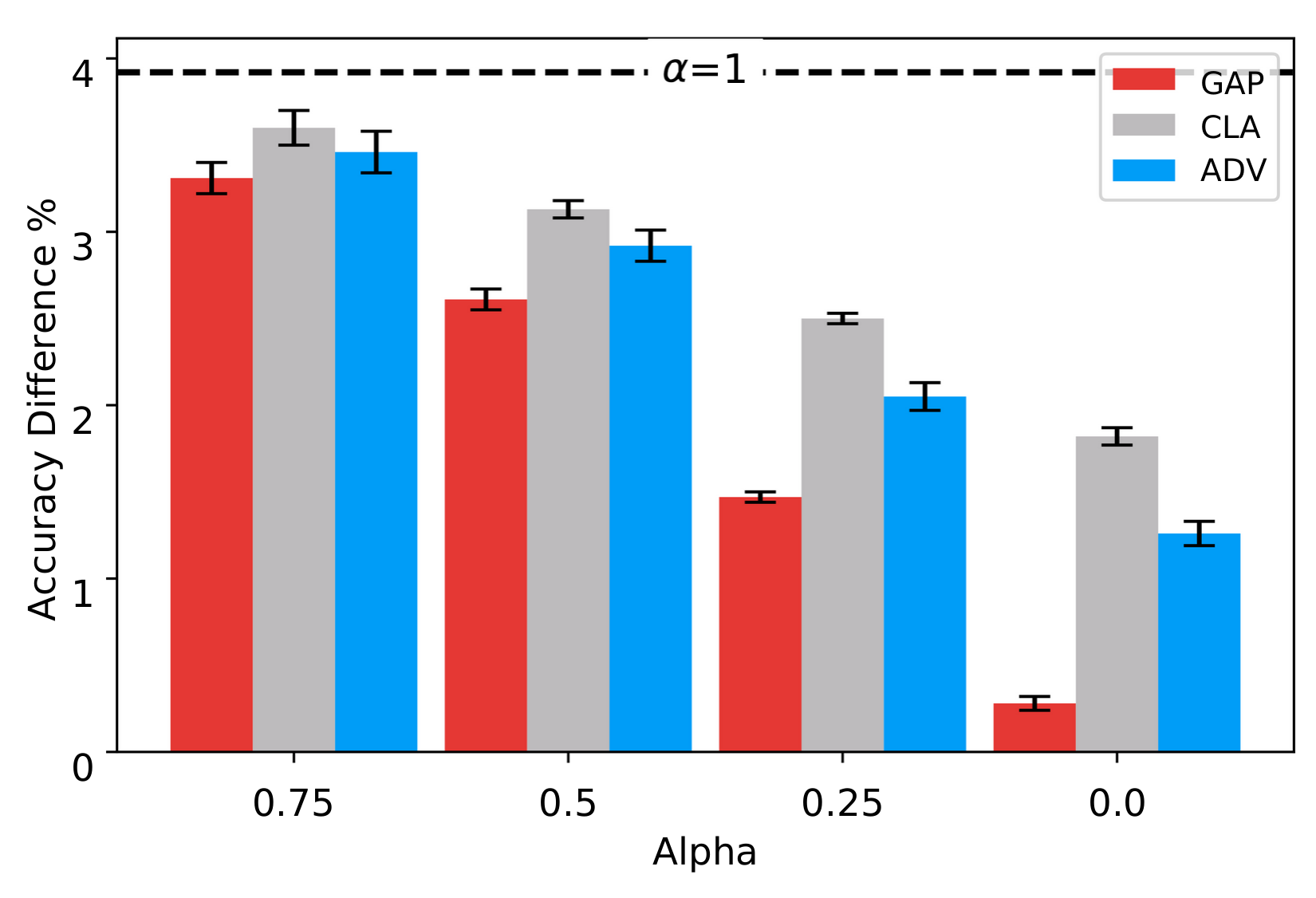}
		\vspace{-1.75em}
		\caption{Wilds: AD}
	\end{subfigure}
	\begin{subfigure}{0.45\linewidth}
		\centering
		\includegraphics[width=\linewidth]{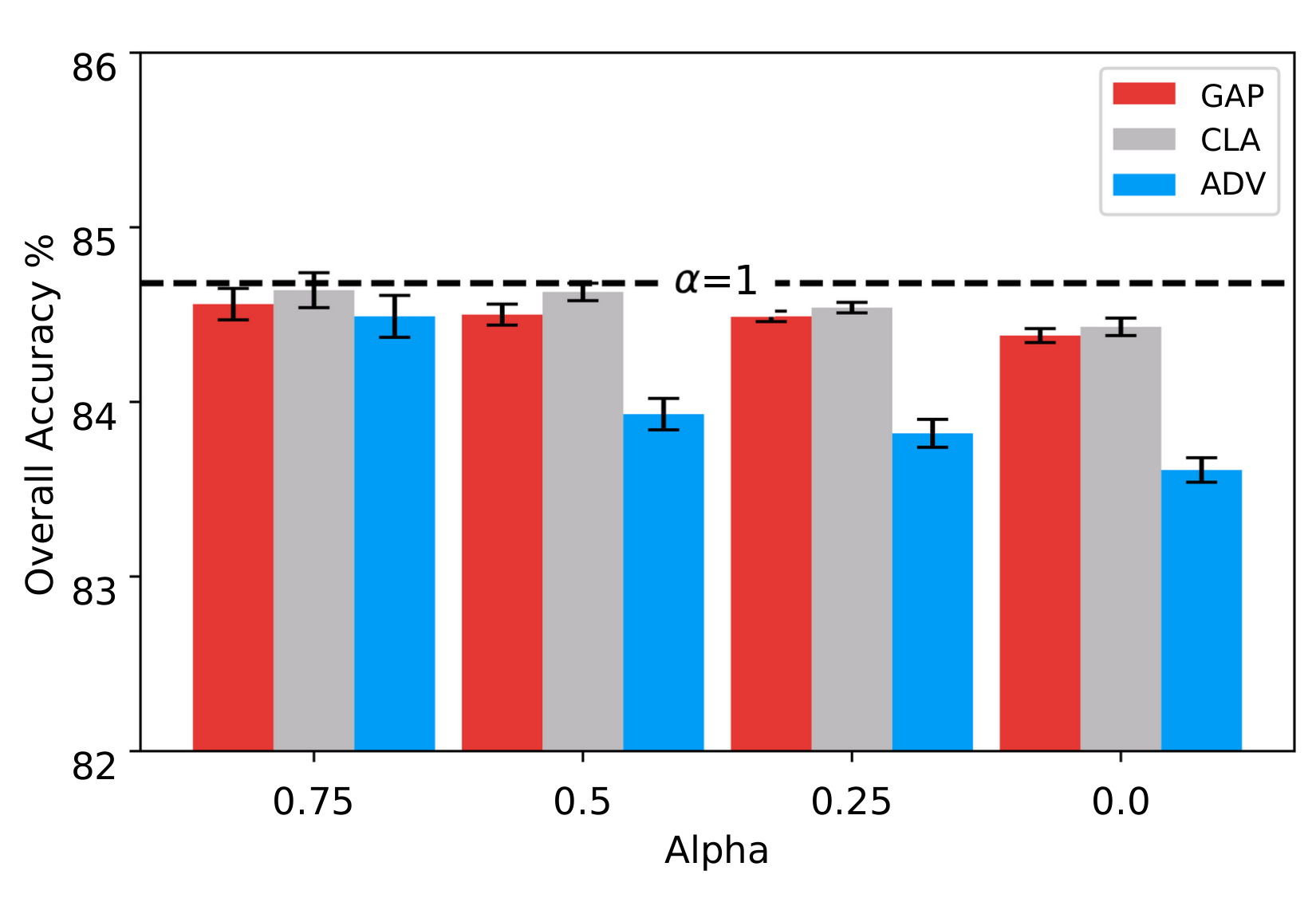}
		\vspace{-1.85em}
		\caption{Wilds: OA}
	\end{subfigure}

% \subfloat[]{%
% 		\includegraphics[width=.24\textwidth]{figs/david_diff.pdf}
% 		\vspace{-0.75em}
% 		\label{fig:davidson_ad}
% }
% \subfloat[]{%
% 		\includegraphics[width=.24\textwidth]{figs/david_acc.pdf}
% 		\vspace{-0.75em}
% 		\label{fig:davidson_oa}
% }
% \subfloat[]{%
% 		\includegraphics[width=.24\textwidth]{figs/jigsaw_diff.pdf}
% 		\vspace{-0.75em}
% 		\label{fig:wilds_ad}
% }
% \subfloat[]{%
% 		\includegraphics[width=.24\textwidth]{figs/jigsaw_acc.pdf}
% 		\vspace{-0.75em}
% 		\label{fig:wilds_oa}
% }
    % \vspace{-1.0em}
	\caption{Trade-offs between Accuracy Difference (AD) and Overall Accuracy (OA), on the BERT based model with SUHNPF acting as hypernetwork for three methods  --- GAP (ours), CLA, and ADV -- across the two datasets for $\alpha \in [0,1]$, with $\alpha=0$ optimizing AD only and $\alpha=1$ optimizing OA only. GAP achieves lower AD consistently across $\alpha$ settings and datasets, while a more modest drop in OA is observed across methods as AD is reduced. 
%	observe a decrease in AD from left to right and almost reaching zero for GAP, at the expense of some drop in OA. Similar trends can be observed for CLA and ADV, however with insignificant drop in AD.
	}
	\label{fig:ad_traj}
	\vspace{-1.0em}
\end{figure}

% \vspace{-0.5em}
\subsection{Multi Objective Optimization (MOO)}

In Section \ref{subs:res-soo} we used GAP, CLA, or ADV to directly optimize fairness. However, the reduced AD comes at the cost of lower OA. In order to find the optimal trade-offs between fairness and accuracy, we use the SUHNPF framework in a MOO experimental setup. We use a BERT-based classifier and three different pairs of objective functions: WCE \vs GAP; WCE \vs CLA; and WCE \vs ADV, 
%Having compared optimization of only GAP, CLA, or ADV in isolation in Section \ref{subs:res-soo}, we next evaluate use of the SUHNPF framework to learn optimal trade-offs of WCE \vs GAP, CLA, and ADV for BERT on Davidson and Wilds datasets. 
learning a linear MOO trade-off between the two competing objectives. 

{\bf Fig. \ref{fig:ad_traj}} shows the results of the MOO experiments. SUHNPF allows us to control how important is each objective (accuracy \vs fairness) by choosing the value of $\alpha$. At $\bm{\alpha=1}$, we optimize only for Accuracy, and at $\bm{\alpha=0}$, only for fairness. We illustrate the different trade-offs at 4 points of the Pareto front ($\alpha$ = 0, 0.25, 0.5, and 0.75). We can observe that with decreasing $\alpha$, both AD and OA decrease. For ADV we can see that the drop in AD is comparable to the drop in OA, which is not an efficient trade-off between accuracy \vs fairness. GAP and CLA maintain a relatively consistent OA, while GAP reduces AD far more than CLA, yielding the best trade-off for each $\alpha$. See \textbf{Appendix \ref{app:diverge}} for discussion on metric divergence and tabulated values in experiments.
We can conclude that GAP is consistently the best metric, across SOO and MOO experimental setups and across different values of $\alpha$ for MOO.

\section{Discussion and Conclusion}

\paragraph{Optimizing Fairness:} Since fairness measures embody different underlying assumptions and statistical choices %\citep{Narayanan18,friedler2021possibility}
, selecting an appropriate fairness metric often depends on the task, use case, and stakeholder priorities. %\citep{friedler2019comparative,bellamy2018ai, hu2018welfare}. 
In this work, we focus on a popular fairness objective of balancing accuracy across different demographic groups, also known as minimizing Accuracy Difference.
%, variously known as {\em equal accuracy} \citep{mitchell2021algorithmic}, {\em equality of accuracy} \citep{heidari2019moral}, {\em overall accuracy equality} \citep{berk2021fairness,wiki_fairness}, {\em accuracy equity} \citep{dieterich2016compas}, or minimizing {\em accuracy difference}  (AD) \citep{das2021fairness}. 
We show that our {\em Group Accuracy Parity} (GAP) measure directly optimizes AD without {\em metric divergence} between loss function \vs evaluation metric. Results show GAP consistently achieves lower AD than prior work with modest loss in OA across datasets.  

\paragraph{MOO and Toxic language detection:} Rather than force the users to settle for any single accuracy or fairness measure, we further adopt SUHNPF, a multi-objective optimization (MOO) framing for joint pursuit of multiple objectives. We learn the full Pareto manifold over competing objectives so that users can view the full space of feasible trade-offs and choose any desired trade-off on the solution manifold, \textit{a posteriori}. We empirically demonstrate that our measure GAP performs better than alternative differentiable fairness objectives in reducing AD. To the best of our knowledge this is the first use of MOO for fair toxic language detection.

\paragraph{Fairness and toxic language detection:} We explore two different aspects of fairness in toxic language detection: 1)~fair moderation of posts written by authors from different demographic groups; and 2) fair protection of different groups targeted by posts. We successfully improved the fairness of the models in both experimental setups, demonstrating the generality of the proposed approach.

\paragraph{Extending GAP to multiple classes and demographic groups} We formulate GAP following the strict definition of AD, which is for two classes and two demographic groups. Fairness literature has discussed heuristics and formulations for extending AD to multi-group and multi-class classification and balancing between multiple groups. As a future work, GAP can be extended based on those hypotheses.

\paragraph{Group identification:} With author demographics in \citep{davidson2017automated}'s dataset, we rely on automatic detection of author dialect, which is noisy. With target group demographics in Wilds \citep{koh2021wilds}, we assume oracle knowledge of target groups from annotation, which would have to be noisily detected in practice. In both cases, therefore, we make simplifying assumptions in this work. Optimizing trade-offs with awareness of noise in detection of demographic groups thus remains another direction for future work. % to improve the generality and realism of our findings.

\paragraph{Dataset debiasing:} Recent studies highlight the risks of annotation bias, be it by annotator guidelines or the annotators themselves. \citep{sap2019risk} and \citep{davidson2019racial} analyze the correlation between race and gold-label of toxicity in several datasets and find a strong association between African American English (AAE) markers and toxicity annotation. Because our work is restricted to balancing accuracy across the sensitive attribute, given the dataset as it is annotated, our results our limited by any such bias present in the data \citep{ludwig2024unraveling}, Addressing such annotation bias thus remains another key direction for future work. 

\paragraph{Generality and scope of this work} We implement GAP and SUHNPF for the task of TL detection and demonstrate promising results - improved fairness and computational efficiency. However, our work can be extended to other tasks, datasets, and neural models in any practical situation where ensuring equal accuracy across different demographic groups is a desired objective. Recently, Kovatchev and Lease \citep{kovatchev-lease-2024-benchmark} demonstrated the significant impact of imbalanced data in popular NLP benchmarks. Our work can help address that challenge.

\section*{Acknowledgments} 
We thank the anonymous reviewers for their valuable feedback. This research was supported in part by Amazon, Wipro, the Knight Foundation, the Micron Foundation, and by Good Systems\footnote{\scriptsize\url{http://goodsystems.utexas.edu/}}, a UT Austin Grand Challenge to develop responsible AI technologies. The statements herein reflect the authors'  opinions only.

\bibliographystyle{ACM-Reference-Format}
\bibliography{References}

\appendix

\section{Relating BCE and GAP Measures}% -- Derivation and Analysis}
\label{app:GAP}

We provide a step by step derivation from BCE to GAP measures and analyze how each of the measures are correlated, to highlight their interplay. Before delving into the measures, we setup the notation and classes to illustrate the relation.

\subsection{Binary Cross Entropy}

Binary Cross Entropy (BCE), as formulated in  \textbf{Eq. \ref{eq:bce}} is typically used as a loss function for optimizing a classifier. Although not a strict one-to-one correspondence, it is generally observed that minimizing the BCE loss leads to maximization of Accuracy. The BCE formulation does not consider imbalance across class frequency, hence might be biased towards the majority class label. It also does not consider the sensitive attributes.

\subsection{Weighted Cross Entropy}

One way to account for the imbalance across toxic and non-toxic labels $(y)$ is Weighted Cross Entropy (WCE), a variation of BCE that re-weights the error for the different classes proportional to their inverse frequency of labels $(y)$. This re-weighting strategy is available in popular packages like SkLearn \citep{scikit-learn} and is discussed in detail by \citep{lin2017focal}. 

 \begin{align}
    BCE &\propto  -(\frac{1}{N} \sum \limits_{N} y \log (\hat{y}) + (1-y) \log (1-\hat{y}))  -(\frac{1}{N} \sum \limits_{N} y \log (\hat{y}) + (1-y) \log (1-\hat{y}))  \label{eq:duplicate}
\end{align}
\vspace{-1em}
 \begin{align}
% 	 	\resizebox{0.42\textwidth}{!}{
		 		WCE = \underset{\textrm{BCE loss for toxic class ($y=1$) with scaling}}{\underbrace{-\frac{Q}{N} \sum \limits_{N} y \log (\hat{y}) + (1-y) \log (1-\hat{y})}} \hspace{2em}
		 		\underset{\textrm{BCE loss for non-toxic class $(y=0)$ with scaling}}{\underbrace{-\frac{P}{N} \sum \limits_{N} y \log (\hat{y}) + (1-y) \log (1-\hat{y})}} 
		 		 \label{eq:wce}
		 	% } 
	 \end{align}
%\vspace{-0.5em}

In {\bf Eq. \ref{eq:duplicate}}  we replicate BCE (Eq. \ref{eq:bce}) terms twice which only introduces a duplication without formulation alteration. To ensure class balancing across toxic and non-toxic classes, we scale each of the duplicate terms \wrt to the sample count (toxic: $P$, non-toxic: $Q$) of the opposite class, while performing summation. When there's no class imbalance \ie $P=Q$, WCE reduces to $2\cdot$BCE, which has the same loss trajectory as BCE. This definition of WCE in \textbf{Eq. \ref{eq:wce}} is differentiable, owing to it's similar form to BCE and shares all the properties of BCE which allows it to be used as a loss for optimizing binary classifiers. WCE attempts to reduce the bias of the majority label due to the inverse sample count scaling, \ie majority and minority classes scaled by their opposite sample counts respectively.

\begin{remark}
Rescaling the majority and minority labels $(y)$ with their inverse frequency only ensures reduced bias towards the majority label. It does not optimize for equal accuracy across both the labels.
\end{remark}

\subsection{WCE \wrt Sensitive Group Attribute}

While WCE accounts for the label imbalance in the dataset, it still does not consider the notion of fairness and the different sub-populations. The core idea behind WCE is that we can ``copy'' the loss function twice and then apply mathematical transformations to it, while maintaining the property of differentiability. We apply that same idea to derive our loss function for fairness. We calculate two separate instances of WCE: $WCE(g=1)$, calculated for data samples of group 1, and $WCE(g=0)$, calculated for data samples of group 0. GAP in essence is the 2-norm difference between the WCE's across each sensitive attribute $s$. The GAP loss function in \textbf{Eq. \ref{eq:GAPad}} obtains a minimum only when both WCE errors match across the binary sensitive attribute. Note that unlike WCE, our measure GAP is defined as the difference between the two loss functions, rather than their weighted sum. Therefore GAP reaches its minimum when the two sub-populations of sensitive group attribute $(s)$ achieve the same accuracy.

\section{Datasets} \label{app:dataset}

We consider two datasets: \citep{davidson2017automated} for author demographics and 
the {\em Civil Comments} \citep{borkan2019nuanced} portion of {\em Wilds}
\citep{koh2021wilds} for target demographics. In each case, we frame the task as a binary classification problem (Toxic \vs non-Toxic, or ``safe'') with binary sensitive attributes (Majority \vs Minority, the under-represented, sensitive attribute).
% \noindent \textbf{Author Demographics Dataset} \citep{davidson2017automated}. We first consider author-oriented fairness: fair moderation of posts written by authors from different demographic groups. The sensitive attribute is \textit{race}, as identified by the dialect of the tweets. Following prior work, in the absence of an oracle for identifying the dialect, we apply an off the shelf \citept{blodgett2017dataset}'s model to automatically detect dialect labels for each tweet as African-American English (AAE) or Standard American English (SAE), representing \textit{Minority} and \textit{Majority} groups, respectively. We acknowledge both that dialect is only a weak surrogate representation of demographic race, and that automatic detection of dialect will naturally incur noise. 
For Davidson, since an explicit train-test split does not exist, we randomly seed the dataset into train-test splits of $90\%-10\%$, following SkLearn's \citep{scikit-learn} stratified sampling to ensure similar proportion of positive and negative tweets across the splits. For Wilds \citep{koh2021wilds} we select tweets where more than 50\% of annotators agreed on the gender of the target, and the toxicity label as well. Note that the annotation for male and female in the dataset is carried out separately, so it is possible that a tweet is targeted both towards male and female. We include such tweets in both portions as independent samples. Such pre-processing has been done across both train and test splits for evaluation purposes.

\section{Setup} \label{app:setup}

Experiments use a Nvidia 2060 RTX Super 8GB GPU, Intel Core i7-9700F 3.0GHz 8-core CPU and 16GB DDR4 memory. We use the Keras \citep{chollet2015} library on a Tensorflow 2.0 backend with Python 3.7 to train the networks in this paper. For optimization, we use AdaMax \citep{kingma2014adam} with parameters (\textit{lr}=0.001) and $1000$ steps per epoch. For each configuration, we did five independent runs to report mean and variance.

\section{Runtime} \label{app:runtime}

The benefit of any Pareto HyperNetwork is to trace out the approximated front of feasible values during the training time, so that uses can extract neural weights corresponding to their desired trade-off values \textit{a posteriori}. In our experiments, for the five trade-off values shown, one can achieve it in two ways.
\begin{enumerate}
    \item Run the Bert model five times, each with different trade-off in the loss function
    \item Run the Bert model one time, with the Pareto HyperNetwork supervising it.
\end{enumerate}

The Bert model ran for $10$ epochs with $\sim 10$ mins per epoch, for a total runtime $\sim 100$ mins. If we run the same configuration for five trade-off, that would equate to $\sim 500$ mins of runtime. Thus, any additional trade-off measure the user desires would cost an extra $\sim 100$ mins each. The SUHNPF Pareto HyperNetwork on the other hand approximated a manifold of trade-off values supervising the Bert model, where the Bert model still takes $\sim 100$ mins with the supervising network taking additional $\sim 60$ mins for manifold approximation. Extracting the weights of the Bert model post-hoc takes an additional $\sim 20$ mins for each trade-off. Therefore, while both the prescribed approaches would roughly yield similar results from optimization of the Bert model, Approach 1 would take $\sim 500$ mins, while Approach 2 would take $\sim 260$ mins, resulting in a $\sim 2\times$ speedup via PFL.

\section{Discussion on Metric Divergence} \label{app:diverge}

\textbf{Table \ref{tab:stas_measure}} reports the the Accuracy Difference (AD) and Overall Accuracy (OA) values achieved for the different trade-off configurations of the Bert model, across three loss measures. This is a tabulated version of \textbf{Fig. 1 (main text)}. Note that for trade-off $\alpha=1$, only OA is maximized, hence none of the losses play any part, thus a common number across three columns, for each dataset. As the trade-off takes into account each of the loss measures, we empirically observe GAP to be performing best \wrt the other measures, since it is being optimized \wrt minimizing AD. 

\begin{table}[h]
 	\centering
 	\vspace{-1em}
 	\resizebox{1.0\linewidth}{!}{%
 		\begin{tabular}{c|lcc|lcc|lcc} \toprule
 			$\alpha$ & \multicolumn{3}{c|}{Accuracy Difference} & \multicolumn{3}{c|}{Overall Accuracy} & \multicolumn{3}{c}{F1} \\ \midrule
 			& GAP (Ours) & CLA & ADV & GAP (Ours) & CLA & ADV & GAP (Ours) & CLA & ADV \\ \midrule
 			\multicolumn{10}{c}{Davidson} \\ \midrule
 			1.00 & \multicolumn{3}{c|}{5.9 $\pm$ 0.1} & \multicolumn{3}{c}{88.9 $\pm$ 0.2} & \multicolumn{3}{c}{0.71 $\pm$ 0.02} \\ \midrule
 			0.75 & 4.2 $\pm$ 0.1 & 5.0 $\pm$ 0.1 & 4.7 $\pm$ 0.1 & 88.5 $\pm$ 0.3 & 88.6 $\pm$ 0.2 & 88.2 $\pm$ 0.4 & 0.70 $\pm$ 0.01 & 0.69 $\pm$ 0.01 & 0.68 $\pm$ 0.00 \\
             0.50 & 2.7 $\pm$ 0.1 & 3.7 $\pm$ 0.1 & 3.2 $\pm$ 0.1 & 88.1 $\pm$ 0.5 & 88.3 $\pm$ 0.5 & 87.4 $\pm$ 0.6 & 0.69 $\pm$ 0.02 & 0.67 $\pm$ 0.01 & 0.65 $\pm$ 0.01 \\
             0.25 & 1.2 $\pm$ 0.1 & 2.4 $\pm$ 0.0 & 2.7 $\pm$ 0.1 & 87.7 $\pm$ 0.2 & 87.9 $\pm$ 0.4 & 86.8 $\pm$ 0.6 & 0.67 $\pm$ 0.01 & 0.65 $\pm$ 0.00 & 0.64 $\pm$ 0.01 \\
             0.00 & 0.1 $\pm$ 0.0 & 0.9 $\pm$ 0.0 & 2.4 $\pm$ 0.1 & 87.3 $\pm$ 0.1 & 87.6 $\pm$ 0.2 & 86.3 $\pm$ 0.4 & 0.66 $\pm$ 0.00 & 0.64 $\pm$ 0.02 & 0.61 $\pm$ 0.01 \\
 			\midrule   
 			\multicolumn{10}{c}{Wilds} \\ \midrule
 			1.00 & \multicolumn{3}{c|}{3.9 $\pm$ 0.2} & \multicolumn{3}{c}{84.7 $\pm$ 0.3} & \multicolumn{3}{c}{0.65 $\pm$ 0.02} \\ \midrule
 			0.75 & 3.3 $\pm$ 0.1 & 3.6 $\pm$ 0.1 & 3.5 $\pm$ 0.1 & 84.6 $\pm$ 0.2 & 84.6 $\pm$ 0.1 & 84.5 $\pm$ 0.3 & 0.63 $\pm$ 0.02 & 0.62 $\pm$ 0.01 & 0.62 $\pm$ 0.02\\
 			0.50 & 2.6 $\pm$ 0.1 & 3.1 $\pm$ 0.1 & 2.9 $\pm$ 0.1 & 84.5 $\pm$ 0.4 & 84.6 $\pm$ 0.6 & 83.9 $\pm$ 0.4 & 0.62 $\pm$ 0.0 & 0.61 $\pm$ 0.01 & 0.60 $\pm$ 0.01 \\
 			0.25 & 1.5 $\pm$ 0.0 & 2.5 $\pm$ 0.0 & 2.0 $\pm$ 0.1 & 84.5 $\pm$ 0.1 & 84.5 $\pm$ 0.2 & 83.8 $\pm$ 0.5 & 0.60 $\pm$ 0.01 & 0.60 $\pm$ 0.01 & 0.57 $\pm$ 0.01 \\
 			0.00 & 0.3 $\pm$ 0.0 & 1.8 $\pm$ 0.1 & 1.3 $\pm$ 0.1 & 84.4 $\pm$ 0.1 & 84.4 $\pm$ 0.1 & 83.6 $\pm$ 0.2 & 0.58 $\pm$ 0.02 & 0.58 $\pm$ 0.01 & 0.55 $\pm$ 0.02 \\
 			\bottomrule            
 		\end{tabular}
 		}
 		% \vspace{-0.5em}
 		\caption{Performance of GAP \vs CLA, ADV across two datasets in terms of Accuracy Difference (AD) and Overall Accuracy (OA). GAP achieves lower AD consistently across $\alpha$ settings and datasets, while a more modest drop in OA is observed across methods. $\alpha=1$ minimizes WCE over labels only, hence same error across the three measures.}
 		\label{tab:stas_measure}
 		\vspace{-2em}
 \end{table}

CLA is designed to optimize for Equal Opportunity \ie False Negative Rate across each group of sensitive attribute $(s)$, follows similar trajectory to GAP. As these measures operate on different sections of the confusion matrix,  optimizing for some values in them leads to better numbers in other parts of the table, since the total number of samples are fixed. ADV, on the other hand, tries to balance False Positive Rate across each sub-population of sensitive attribute $(g)$. The performance of ADV however deviates a lot from the trajectory of both GAP and CLA, since their adversarial setup is not strictly optimizing for FPR, and similar deviations can be seen in their original work \citep{xia2020demoting} as well.

There are various ways to define fairness and over 80 \citep{bellamy2018ai} different post-hoc measures for fairness, corresponding to different use-cases. We obtained the best results when using GAP: a measure designed specifically for achieving Accuracy Parity (AP). Other fairness measures such as CLA and ADV can improve the OAE to a certain degree, but are nowhere near as efficient as GAP. Because no fairness measure is universal \citep{Narayanan18}, it is important to pick a loss function that corresponds to the intended fairness goal.

\vspace{-1em}
\section{Performance of Models on Wilds} \label{app:modelwilds}

\textbf{Table \ref{tab:baseline-wilds}} shows the baseline results on the Wilds \citep{koh2021wilds} dataset. The performance of the classifiers are similar \wrt Table \ref{tab:baseline-david}, where due to focus on Overall Accuracy (OA), there is a gap between the group specific accuracies. This shows the existing bias across the three neural models, with the BERT based model performing relatively better than the rest.

\begin{table}[h]
	\centering
	\vspace{-1em}
	% \resizebox{\linewidth}{!}{%
		\begin{tabular}{l|c|c|c|c}
			\toprule
			Models & Overall \% & Majority \% & Minority \% & AD \% \\ \midrule
			CNN & 83.90 $\pm$ 0.2 & 86.11 $\pm$ 0.1 & 81.27 $\pm$ 0.2 & 4.84 $\pm$ 0.2 \\
			BiLSTM & 83.94 $\pm$ 0.1 & 85.98 $\pm$ 0.2 & 81.52 $\pm$ 0.2 & 4.46 $\pm$ 0.1 \\
			BERT & 84.71 $\pm$ 0.3 & 86.53 $\pm$ 0.1 & 82.49 $\pm$ 0.2 & 4.04 $\pm$ 0.2 \\ \bottomrule
	\end{tabular}
 % }
	% \vspace{-0.5em}
	\caption{Baseline accuracy results on Wilds \citep{koh2021wilds} dataset 
	when maximizing overall accuracy (OA) only. Results show consistent bias of higher accuracy for the Majority.}
	\vspace{-3em}
	\label{tab:baseline-wilds}
\end{table}

\end{document}